\documentclass[aps,pre,floatfix,showpacs,nofootinbib]{revtex4}
\usepackage{graphicx}
\usepackage{epstopdf}
\usepackage{amsmath} \usepackage{amssymb}
\newcommand{\comment}[1]{}
\newcommand{\nunu}{{\boldsymbol \nu}}
\newcommand{\ff}{{\boldsymbol \theta}}

\newcommand{\BEQ}{\begin{equation}}
\newcommand{\EEQ}{\end{equation}}
\newcommand{\BEA}{\begin{eqnarray}}
\newcommand{\EEA}{\end{eqnarray}}
\renewcommand{\d}{{\rm d}}
\renewcommand{\u}{\underline}
\newcommand{\w}{\widetilde}
\newcommand{\ve}{\varepsilon}
\newcommand{\si}{\sigma}

\newcommand{\zmu}{\mu}

\begin{document}
\draft
\title{Relating Zipf's law to textual information}

\author{ Weibing Deng$^{1)}$ and Armen E. Allahverdyan$^{2),}$ 
\footnote{wdeng@mail.ccnu.edu.cn, armen.allahverdyan@gmail.com}}

\address{ $^{1)}$Key Laboratory of Quark and Lepton Physics (MOE) and Institute of Particle Physics,  Central China Normal University, Wuhan 430079, China,\\
  $^{2)}$Yerevan Physics Institute, Alikhanian Brothers Street 2,
  Yerevan 375036, Armenia}


\begin{abstract}

Zipf's law is the main regularity of quantitative linguistics. Despite
of many works devoted to foundations of this law, it is still unclear
whether it is only a statistical regularity, or it has deeper relations
with information-carrying structures of the text.  This question relates
to that of distinguishing a meaningful text (written in an unknown
system) from a meaningless set of symbols that mimics statistical
features of a text. Here we contribute to resolving these questions by
comparing features of the first half of a text (from the beginning to
the middle) to its second half. This comparison can uncover hidden
effects, because the halves have the same values of many parameters
(style, genre, author's vocabulary {\it etc}). In all studied texts we
saw that for the first half Zipf's law applies from smaller ranks than
in the second half, i.e. the law applies better to the first half. Also,
words that hold Zipf's law in the first half are distributed more
homogeneously over the text. These features do allow to distinguish a
meaningful text from a random sequence of words. Our findings correlate
with a number of textual characteristics that hold in most cases we
studied: the first half is lexically richer, has longer and less
repetitive words, more and shorter sentences, more punctuation signs and
more paragraphs. These differences between the halves indicate on a
higher hierarchic level of text organization that so far went unnoticed
in text linguistics. They relate the validity of Zipf's law to textual
information. A complete description of this effect
requires new models, though one existing model can account for some of
its aspects. 

\comment{
Zipf's law is the main regularity of quantitative linguistics.
Despite of many works devoted to foundations of this law, it is still
unclear whether it is only a statistical regularity, or it has deeper
relations with information-carrying structures of the text. Here we
contribute to resolving these questions by comparing features of the
first half of a text to its second half. This comparison can uncover
hidden effects, because the halves have the same values of many
parameters (style, genre {\it etc}). We found that for the first half
Zipf's law applies from smaller ranks than in the second half. Also,
words that hold Zipf's law in the first half are distributed more
homogeneously over the text. Our findings correlate with a number of
textual characteristics that hold in most cases we studied: the first
half is lexically richer, has longer and less repetitive words, more and
shorter sentences, more punctuation signs and more paragraphs. These
differences between the halves indicate on a higher hierarchic level of
text organization that so far went unnoticed in text linguistics. They
relate the validity of Zipf's law to textual information. Some of these
results are described by a statistical physics model. 
}

\end{abstract}
\maketitle

\section{Introduction}

\label{intro}

Quantitative and universally applicable relations are rare in social
sciences. Each such relation acquires the status of law and paves the
way towards bringing in methods and ideas of natural sciences. This is
why Zipf's law|discovered independently by stenographer
Estoup in 1912 \cite{estoup} and physicist Condon in 1928 \cite{condon},
and later on advertized by linguist Zipf
\cite{zipf,joos,wyllis,baa}|attracted so much inter-disciplinary
attention. The law applies both to text mixtures (corpora), and to
separate texts written in many natural and artificial alphabetic
languages \cite{greek,indian,moreno}, as well as in Chinese characters
\cite{epjb}. It states that in a given text the ordered and normalized
frequencies $f_1>f_2> ...$ for the occurrence of the word with rank $r$
hold $f_r\propto r^{-\gamma}$ with $\gamma\approx 1$ \cite{wyllis,baa}. 

Rank-frequency relations imply coarse-graining, e.g. since they are
invariant with respect to permutation of words. This is one reason why
there are many approaches towards deriving this law, but none of them is
conclusive about its origin. Existing approaches can be roughly divided
into two groups: {\it (i)} optimization principles
\cite{shreider_sharov,sole,prokopenko,dickman,mandelbrot,dunaev,mitra,manin,mandel,arapov,shrejder,dover,vakarin,liu,baek};
{\it (ii)} statistical approaches
\cite{li,simon,zane,kanter,hill,pre,latham}. 

Note that {\it (i)} includes Zipf's program|which is so far not
conclusive \cite{prokopenko,dickman}|that the language trades-off
between maximizing the information transfer and minimizing the
speaking-hearing effort. The law can be also derived from various
generalizations of the maximum entropy method
\cite{mandel,arapov,shrejder,dover,vakarin,liu,baek}, though the choice
of the entropy function to be maximized (and of relevant 
is neither unique nor clear. The general problem of derivations from
{\it (i)} is that verifying the law for a frequency dictionary (or for a
large corpus) does not yet mean to explain it for a concrete text. 
E.g. if the word frequencies obeying Zipf's law are
deduced from considerations related to the meaning of words
\cite{manin}, then the applicability to a single text is unclear, since
the fact that the words normally have widely different frequencies in
different texts requires a substantial reconsideration of the word's
meaning in each text (this is not the case in real texts). 

The first model within {\it (ii)} was a random text, where words are
generated through random combinations of letters, i.e. the most
primitive stochastic process \cite{mandelbrot,li}.  Its drawbacks
\cite{howes,seb,cancho} (e.g. many words having the same frequency) are
avoided by more refined models \cite{simon,zane,kanter,hill}, though
they also do not explain the region of rare words (hapax legomena). For
instance, the text growth model studied in \cite{simon} was later on
shown to fail in describing the hapax legomena \cite{minn}. Note that
the Zipf's law cannot explianed via fixed probability of words (e.g.
estimated from those of a frequency dictionary), since the same word can
have widely different probabilities in different texts that obey Zipf's
law \cite{arapov}. This drawback is absent in a recent probability model
that is based on latent variables and deduces Zipf's law together with
its applicability to a single text and its extension to rare words
\cite{pre}.

Despite of (or even due to) these efforts, there is a major open question
\cite{orlov,cancho,pian}: is Zipf's law {\it only} a statistical
regularity, or it {\it also} reflects information-carrying structures of
a meaningful text? This question relates to one of fundamental issues in
linguistics: how a text written in an unknown system (e.g. the Voynich
manuscript) can be efficiently distinguished from a meaningless
collection of words \cite{baa}. 

Here we contribute to resolving these questions by noting that natural
texts evolve from beginning to end. This obviously important notion is
absent from the rank-frequency relation, which is invariant with respect
to any permutation of words.  Thus we divide texts into halves, each
one containing the same amount of words. This implies a semantic
difference: the first half can be understood without the second one, but
normally the second half is not easy to understand without the first
half. The first part of the text normally contains the exposition (which
sometimes can be up to 20 \% of the text), where the background
information about events, settings, and characters is introduced to
readers. The first part also plots the main conflict (open issue), whose
denouement (solution) comes in the second half
\footnote{\label{foo1}Scientific texts contain closely related aspects:
introduction, critique of existing approaches, statement of the problem,
resolution of the problem, implications of the resolution {\it etc}. The
discussion on differences between the halves applies also here.}. 

Dividing the text into two halves neutralizes confound variables that
are involved in a complex text-producing process (style, genre, subject, the
author's motives and vocabulary {\it etc}), because they are the same in
both halves. Hence by comparing the two halves with each other we hope to
see regularities that are normally shielded by above variables. In all
texts we studied we noted the following regularities.

({\it 1}) For the first half Zipf's law applies from smaller ranks than
in the second half, i.e. the law applies better to the first half, than
to the second half. The smallest rank is the major limiting factor in
applicability of the law, as shown by section \ref{divi}. 

({\it 2}) For the first half the words that hold the law are distributed
more homogeneously (in the properly quantified sense) along the text;
see section \ref{spatial}. 

These features are specific for meaningful texts and they can be
employed for distinguishing meaningful texts from a random collection of
words that happens to hold Zipf's law, e.g. due to one of numerous
stochastic mechanisms reviewed above
\cite{mandelbrot,li,mandel,arapov,shrejder,dover,vakarin,liu,baek,latham,pre}. 

We related the above results on Zipf's law to textual information.
Rendering more detailed discussion till section \ref{textual}, we note
that meaningful texts consists of several hierarchic levels: words,
phrases, sentences (clauses), paragraphs \cite{hutchins,valgina,hasan}.
We looked whether the first and second halves differ with respect to
quantitative characteristics of these levels. We identified several such
differences:

({\it 3}) The first half is lexically richer (contains more distinct
words and more rare words), has longer and less repetitive words,
shorter sentences, more punctuation signs and more paragraphs. 

In contrast to ({\it 1}) and ({\it 2}), some features within ({\it 3})
hold in most cases, but not strictly in all cases we studied. Despite of
such minor exclusions, the results are suggestive in pointing out that
the validity domain of Zipf's law relates to features of a
meaningful text. 

This paper is organized as follows. The next section discusses our
method of studying Zipf's law and its validity range. In particular, 
we explain how the validity range of Zipf's law behaves under mixing 
(joining together) two or more texts; see section \ref{mixing}. Section
\ref{divi} compares the halves of a text with respect to the validity
range of Zipf's law and the amount of rare words.  These results are
illustrated on Fig.~\ref{f1}. Section \ref{textual} reminds several
aspects of textual information known in linguistics, designs on their
base several straightforward quantitative characteristics and checks
them for two halves. The results are summarized in Table~\ref{tab0}.
Section \ref{spatial} studies the distribution of words along the text.
Here we study this distribution for different halves and relate it with
the validity range of Zipf's law; see Fig.~\ref{f2}.  Section
\ref{theor} applies the theory for Zipf's law and rare words
proposed in \cite{pre} for describing several aspects of our findings.
Here we emphasize that this theory is incomplete. Better theories are
yet to be found. We summarize our results in the last section.

\section{Phenomenology of Zipf's law and its validity range}
\label{validity}

\subsection{The method of searching for Zipf's law}
\label{choban}

\begin{figure} \centering
\includegraphics[width=0.8\textwidth]{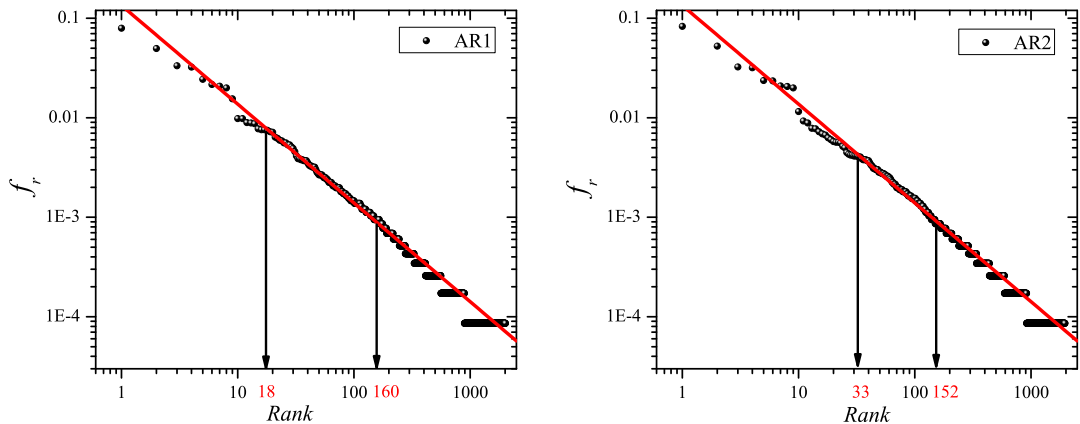} 
\caption{ (Color online) Frequency $f_r$ vs. rank for the first (AR1)
and second (AR2) halves of the text {\it The Age of Reason} (AR) by T.
Paine, 1794 (the major source of British deism); see (\ref{w1}) for the
definition of $f_r$. Red lines show the Zipf curves.  Arrows indicate on
$r_{\rm min}$ and $r_{\rm max}$, i.e. on the validity range of the
Zipf's law. For this example, the power-law exponent $\gamma$ for both
halves are equal to $0.99$, while the constant $c$ for both halves are
equal $c_1=c_2=0.135$; see (\ref{goodwin}) and Table~\ref{tab1} in
Appendix I. The step-wise behavior of $f_r$ for $r>r_{\rm max}$ refers
to hapax legomena. } \vspace{0.25cm} \label{f1} \end{figure}

We explain how we recover Zipf's law from the data; see 
Appendix II and \cite{epjb,pre} for details. For a given text we
extract the ordered frequencies of different words
\footnote{\label{janmuller}There is no universal definition of word
\cite{muller}; e.g. there is a natural uncertainty on whether to count
plurals and singulars as different words. Different definitions of word
can produce numerically different results \cite{muller}. We mostly work
with methods that assume singular and plural to be different words. But
we also checked that our qualitative conclusions do hold as well when
singular and plural are taken to be the same word.} ($n$ and $N$ are
respectively the number of different words and the overall number of
words in a text):
\BEA 
\label{w1} \{f_r\}_{r=1}^{n}, \qquad f_1\geq ...\geq f_{n}, \qquad
{\sum}_{r=1}^{n} f_r =1. \EEA 
The data $\{f_r\}_{r=1}^{n}$ is fit to a power law: 
\BEA
\label{barbos}
\hat{f}_r=cr^{-\gamma}. 
\EEA
Fitting parameters $\gamma$ and $c$ were found from minimizing the
sum of squared errors: $S_{\rm err}$; see Appendix II. The fitting
quality is found from by the minimized value of $S_{\rm err}^*$ and from the
coefficient of determination $R^2$, which is the amount of variation
in the data explained by the fitting; see Appendix II. Hence $S_{\rm
err}^*\to 0$ and $R^2\to 1$ mean good fitting.  We minimize $S_{\rm
err}$ over $c$ and $\gamma$ for $r_{\rm min}\leq r\leq r_{\rm max}$ and
find the minimal $r_{\rm min}$ and the maximal $r_{\rm max}$ for which
\begin{equation}
\label{dard}
S_{\rm err}^*\leq 0.05 ~~{\rm and}~~ 1-R^2\leq 0.005. 
\end{equation}
These values of $r_{\rm max}$ and $r_{\rm min}$ also determine the final
fitted values $c^*$ and $\gamma^*$ of $c$ and $\gamma$, respectively;
see Fig.~\ref{f1} and Tables \ref{tab1}, \ref{tab2}. Thus $c^*$ and
$\gamma^*$ are found simultaneously with the validity range $[r_{\rm
min},r_{\rm max}]$ of the law. For simplicity, we refer to $c^*$ and
$\gamma^*$ as $c$ and $\gamma$, respectively. The fitting
quality was confirmed via the Kolmogorov-Smirnov (KS) test; see Appendix
III. 

The above method is standard, it differs from others by more rigorous
criteria (\ref{dard}), and by explicitly accounting for the validity
range $r_{\rm min}\leq r\leq r_{\rm max}$ of the power law
(\ref{barbos}). The general idea of fitting ranked frequencies
(\ref{w1}) to the power-law (\ref{barbos}) can be rightly criticized on
the ground that the definition of rank is not independent from the
frequency, hence the small value of $S_{\rm err}^*$ in (\ref{dard}) does
not ensure against correlated errors of rank and frequency \cite{pian}.
We stress that the several aspects the above fitting results will be
recovered in section \ref{spatial}, where the frequency will be
given an alternative representation. 

\subsection{Single texts}

\subsubsection{Parameters of Zipf's law (\ref{barbos}): $\gamma $ and $c$}

We studied 10 English texts written in different epochs and on different
subjects; see Appendix I for their description. We were not able to
increase the number of studied texts substantially (e.g. ten times),
because for each single texts we studied many different features; see
Table~I. Studying them for 100 text will be time-consuming. 

For all studied texts we obtain [see Fig.~\ref{f1}] \BEA \label{goodwin}
\gamma\approx 1, \EEA The exponent $\gamma\approx 1$ came itself, since
we only imposed the power law in (\ref{barbos}). As for the magnitude of
$c$ in (\ref{barbos}), we have two constraints. First note from
Fig.~\ref{f1} that apart of minor exclusions, (\ref{barbos}) is an upper
bound for the frequencies at all ranks. This holds more generally
\cite{wyllis,baa,pre}, and implies a lower bound on $c$:
$c\sum_{k=1}^nk^{-1}>1$. Second, we note that for the obvious constraint
for $r_{\rm min}<r<r_{\rm max}$, the power-law (\ref{barbos}) is very
close to observed frequencies. Hence we have $c\sum_{k=r_{\rm
min}}^{r_{max}}k^{-1}<1$, which leads to an upper bound for $c$. These
bounds are consistent with actual values of $c$, which for the studied
texts hold $0.1<c< 0.2$; see Tables \ref{tab33} and \ref{tab2}. 

\subsubsection{Minimal rank $r_{\rm min}$}

Fig.~\ref{f1} shows that words with ranks $1\leq r<r_{\rm min}$ do not
hold Zipf's law (\ref{barbos}). Among them, the most frequent 3-4 words
relate to the author, since their frequencies coincide for both halves
of the text; see Fig.~\ref{f1}. But other words in the range are
different for both halves; they do not hold Zipf's law due to their
irregular behavior. Table III in Appendix I lists the values of $r_{\rm
min}$ for various texts. 

The range $1\leq r<r_{\rm min}$ range contains mainly function words.
They serve for establishing grammatical constructions (e.g., {\it the,
and a, such, this, that, where, were}) \footnote{Functional words do
have meaning, but it is a general one, e.g. {\it and} refers to joining
and unification, while {\it but} to exclusion. }.  The majority of words
in the Zipfian range $r_{\rm min}<r<r_{\rm max}$ do have a narrow
meaning (content words). A subset of content words has a meaning that is
specific for the text, i.e. they are key-words of this text. The fact
that key-words are located in the Zipfian range was employed for
automatic indexing of texts \cite{ibm}. Few keywords appear also in the
range $1\leq r<r_{\rm min}$, e.g.  {\it love} and {\it miss} for the
romance novella DL and {\it god} and {\it man} for the theological AR;
see Table~\ref{tab1}. Some keywords are also located in the range
$r>r_{\rm max}$, e.g. {\it eloi} for the science fiction TM, but the
majority of them are in the range $r_{\rm min}<r<r_{\rm max}$. 

We stress that the applicability of Zipf's law to a single text
cannot be explained via text-independent probabilities of words, because
even if the same word enters into different texts it typically has quite
different frequencies there \cite{arapov}, e.g.  among 83 common words
in the Zipfian ranges of texts AR and DL [see Table~\ref{tab1}], only 12
words have approximately equal ranks and frequencies. 

\subsubsection{Maximal rank $r_{\rm max}$}

Now $r_{\rm max}$ is the maximal rank, where (\ref{barbos}) holds
according to criteria (\ref{dard}); see Table~\ref{tab33} for examples.
It appears that $r_{\rm max}$ is in a sense the largest possible rank,
because for $r> r_{\rm max}$ no smooth rank-frequency relation
(including (\ref{barbos})) is expected to work due to words having the
same frequency. Put differently, Zipf's law cannot hold for $r>r_{\rm
max}$, because now the rank-frequency relation consists of steps: many
words having the same frequency $f_r$ \cite{baa}; see Fig.~\ref{f1}. 

The empirical value of $r_{\rm max}$ appeared to be such
that the number of words having frequency $f_{r_{\rm max}}$ is $\lesssim
10$. The absolute majority of different words with ranks in $[r_{\rm
min}, r_{\rm max}]$ have different frequencies; see Fig.~\ref{f1}. It is
only at the vicinity of $r_{\rm max}$ that words having the same start
to appear.  Now the quality of fitting (defined by (\ref{dard})) does
not dependend on small changes of $r_{\rm max}$, but it {\it is}
sensitive to small changes of $r_{\rm min}$. Hence for simplicity we
fixed $r_{\rm max}$ such that the number of words with frequency
$f_{r_{\rm max}}$ is $10$. 

We thus emphasize that the status of $r_{\rm max}$ is different from
$r_{\rm min}$, though they both determine the applicability domain of
the smooth rank-frequency (\ref{barbos}).

\subsubsection{Hapax legomena}

In rank-frequency relation, a sizable number of words appear only very
few times ({\it hapax legomena}).  These rare words amount to a finite
fraction of $n$ (i.e. the number of different words).  The existence and
the (large) number of rare events is not peculiar for texts, since there
are statistical distributions that can generate samples with a large
number of rare events \cite{baa}; see section \ref{theor}.  One reason
why many rare words should appear in a meaningful text is that a typical
sentence contains functional words (which come from a small pool), but
it also has to contain some rare words, which then necessarily have to
come from a large pool \cite{latham} \footnote{\label{lato}E.g. this sentence
contains rare word {\it typical} and {\it pool} that in the present text
are met only 3 and 2 times, respectively.  It also contains frequent
words {\it words}, {\it since}, {\it large}.}.  Though rare words cannot
be described by a smooth rank-frequency relation (including Zipf's
law), their distribution is closely related to the proper Zipf's law
\cite{pre}; see section \ref{theor}.

\subsection{Mixing of several texts}
\label{mixing}

Mixing (joining together) two texts is a standard procedure in
quantitative linguistics. Most of our knowledge on rank-frequency
relations is verified on corpora, i.e. large mixtures of many texts; see
\cite{willi} for a recent discussion.  We shall now follow in detail how
$r_{\rm max}$ and $r_{\rm min}$ behave under mixing. 

\comment{
If Zipf's law is a statistical regularity, then both its validity range
and its precision should increase when mixing several texts, i.e. when
going from a single to several texts together. We can measure the
validity range via the length $r_{\rm max}-r_{\rm min}$ of the Zipfian
range. }

Table~\ref{tab33} shows that the validity range $r_{\rm max}-r_{\rm
min}$ of Zipf's law (\ref{barbos}) clearly increases upon mixing different text: 
\BEA
\label{moh}
r_{\rm max}[A+B]-r_{\rm min}[A+B]
\geq {\rm max}\left(\,
r_{\rm max}[A]-r_{\rm min}[A], \, r_{\rm max}[B]-r_{\rm min}[B]
\, \right),
\EEA
where $A$ and $B$ are different texts, and $A+B$ means the text got by
mixing them. The main reason for increasing 
$r_{\rm max}-r_{\rm min}$ is that the number of different words raises upon 
mixing two different texts, and then $r_{\rm max}$ raises as well, because it
is determined by a condition $f_{r_{\rm max}}\lesssim 10$, as we saw above. 
Hence $r_{\rm max}$ also increases sizably [see Table~\ref{tab33}]:
\BEA
\label{abu}
r_{\rm max}[A+B]\geq {\rm max}\left(\,
r_{\rm max}[A], \, r_{\rm max}[B]
\, \right).
\EEA
Eqs.~(\ref{moh}, \ref{abu}) are expected if Zipf's law is 
a statistical regularity. They ensure the applicability 
of Zipf's law to corpora, where (\ref{barbos}) applies to most of ranks \footnote{In the
context of text-mixing, we note that Ref.~\cite{willi} that mixing
together large corpora brings in|for sufficiently large ranks $r>r_1$|an
additional scaling regime $f_r\simeq r^{-\gamma_1}$ that holds for the
frequency $f_r$ versus rank $r>r_1$ with $\gamma_1$ sizably different
from $1$. This second regime is naturally limited by ranks, where the
rare words appear (hapax legomena). The Zipfian regime $f_r\simeq
r^{-1}$ is thus confined to sufficiently small ranks $r>r_1$. We note
that this new scaling regime emerges only for mixing of large corpora
from different authors and from different topics, whose length exceeds
those of an average text. Hence the second critical regime is not
relevant for single texts that are at focus of our investigation. 
This situation is somewhat similar to Chinese texts \cite{epjb}, where 
sufficiently small texts hold Zipf's law, but mixtures of already 
several texts do not hold this law for large ranks, but before the 
regime of rare words sets in \cite{epjb}.}. 

However, the behavior of $r_{\rm min}$ is less expected [see Table~\ref{tab33}]:
\BEA
\label{bakr}
r_{\rm min}[A+B]\geq {\rm min}\left(\,
r_{\rm min}[A], \, r_{\rm min}[B]
\, \right).
\EEA
Hence for certain texts $r_{\rm min}$ can increase under mixing, i.e.
limit the applicability of the Zipf's law for small ranks. Whether
$r_{\rm min}$ will increase or decrease under mixing depends on the
texts.  At any rate, the change of $r_{\rm min}$ does not have any
significant influence on the increase of $r_{\rm max}-r_{\rm min}$ in
(\ref{moh}). 

Let us now return to the precision of Zipf's law. We applied strict criteria
(\ref{dard}) to obtaining its validity range.  One can also look for
weaker precision measures, e.g.  $d=\sum_{k=r_{\rm min}}^{r_{\rm max}}
(ck^{-\gamma}-f_k)$ that measures how the overall frequency of the
Zipfian range is approximated by Zipf's law; cf.~(\ref{w1},
\ref{barbos}). Table~\ref{tab33} shows that $|d|$ is sufficiently small
so that the applicability of the law is warranted.  But $|d|$ can both
increase and decrease upon mixing two texts; see Table~\ref{tab33}.

Summarizing all the arguments, we can say that overall the validity
range of Zipf's law tends to increase under mixing.  Hence the hope of
Refs.~\cite{orlov,arapov} that Zipf's law applies more precisely to a
single text than to text mixtures do not hold. This conclusion is not
an automatic consequence of Zipf's law, and is specific for alphabetic
writing systems; e.g.  relatively short texts written in Chinese
characters do hold Zipf's law in the above sense, but their mixtures do
not \cite{epjb}.  But at least for alphabetical texts, the law holds
also for text corpora, and hence reflects statistical regularities. The
question is whether it reflects {\it only} statistical regularities.
Mixing is not appropriate for answering this question.  Below we shall
do the opposite, i.e. divide a text into two halves. 


\begin{table} 
\centering
\begin{tabular}{l|c|c} \hline\hline
 &  First half & Second half \\
\hline
Minimal rank of Zipf's law $\,^*$; see section \ref{choban} & -- & + \\
\hline
Maximal frequency of the law; see (\ref{mono}) & + & -- \\
\hline
Spatial homogeneuity of words that hold Zipf's law  $\,^*$ & + & --  \\
\hline
Number of different words & + & -- \\
\hline
Number of rare words (absolute and relative) $\,^*$ & + & -- \\ 
\hline
Normalized prefactor $c/n$ of Zipf's law $\,^*$; see (\ref{bri})   & -- & + \\
\hline
Repetitiveness of words (Yule's constant); see (\ref{yule}) & -- & + \\
\hline\hline
Number of punctuation signs & + & -- \\
\hline
Number of letters & + & -- \\
\hline
Average length of words & + & -- \\
\hline
Number of sentences & + & -- \\
\hline
Average length of sentence & -- & + \\
\hline
Entropy and variance of sentence length distribution in words   & -- & + \\
\hline
Number of paragraphs & + & -- \\
\hline
Size in bytes  & + & -- \\
\hline\hline
Compressibility of the size & $\emptyset $ & $\emptyset $ \\
\hline
Number of functional words  & $\emptyset $ & $\emptyset $ \\
\hline
Exponent of Zipf's law  & $\emptyset $ & $\emptyset $ \\
\hline
Maximal rank of Zipf's law  & $\emptyset $ & $\emptyset $ \\
\hline
\end{tabular}
\caption{\label{tab0}Qualitative comparison of
various features of texts between first and second halves: + (--) means
that the feature is larger (smaller) in the corresponding half.
$\emptyset $ means that the sought difference does not show up. $\,^*$
means that the regularity did not show any exclusion (for other features
we tolerate one exclusion per 10 texts). Features are divided into three
groups by double-lines. The first group does not require anything
beyond separation of a text into different words. Features from the
second group indicate that the first half has more thematic information
than the second half. } 
\end{table}


\comment{
$\bullet$ The minimal frequency of the Zipfian domain holds
\begin{eqnarray}
  \label{boundary1}
  f_{r_{\rm max}} > {c}/{n}.
\end{eqnarray}
We checked that this is valid not only for separate texts but also
for the frequency dictionaries of English and Irish.
For our texts a stronger|but less precise|relation holds
\begin{eqnarray}
  \label{boundary2}
f_{r_{\rm max}} \gtrsim \frac{1}{n}.
\end{eqnarray}
Hence $f_{r_{\rm max}}N \gtrsim \frac{N}{n}\gg 1$; see Table I.
}

\section{Dividing the text into two halves}
\label{divi}

\subsection{Validity range of Zipf's law for each half}
\label{dividivi}

We divided the studied texts into two halves along the flow of the
narrative, i.e. from the beginning to end. Several aspects of the text
are left unchanged, e.g. they are still sufficiently large for
statistics to apply, they have the same overall number of words, the
same author, genre {\it etc}. They are different semantically,
since the first half can be understood without the second half, but the
second half generally cannot be understood alone. Also, the structure of
narrative is different: the first half normally contains the exposition,
where actors, situations and conflicts are set and defined, while the
second half normally contains the denouement; cf.~Footnote~\ref{foo1}. 

Our first observation is that in {\it all} texts we studied the rank
$r_{\rm min}$|where Zipf's law starts|is smaller for the first half,
than the second half [see Table~\ref{tab0} for a qualitative summary of
our results, and Tables~\ref{tab1}, \ref{tab33} and \ref{tab2} for
numeric data]: \BEA \label{baruch} r_{\rm min\,1}< r_{\rm min\,2}.  \EEA
We recall that the fitting quality of Zipf's law depends strongly on
$r_{\rm min}$. It does not depend much on $r_{\rm max}$, because the
latter approximately coincides with a rank, where any smooth
rank-frequency relation will stop to hold due to many words with the
same frequency. Eq.~(\ref{baruch}) is consistent with
(\ref{bakr}), if the two halves [$A$ and $B$ in (\ref{bakr})] are
regarded as different texts that produce the full text when taken
together. There is another relation closely related to (\ref{baruch})
[see Table~\ref{tab2}]
\BEA
\label{mono}
c_1/r_{\rm min\,1}>c_2/r_{\rm min\,2},
\EEA
which shows that in the first half Zipf's law applies from larger
frequency values than in the second half; cf.~(\ref{barbos}, \ref{goodwin}). Inequality
(\ref{mono}) holds for all studied texts apart of one exception in 10
texts. We tolerate such exceptions taking into account various
subjective factors that influence the text formation process. 

Recall that the words with ranks $1\leq r<r_{\rm min}$ are of two
types; cf. the second paragraph after (\ref{goodwin}). The few most
frequent ones are author-specific, since they have the same frequency in both
halves; see Fig.~\ref{f1}. The remaining words with ranks from
$1<r<r_{\rm min}$ are not author-specific. Their amount is smaller in
the first half of the text compared to the second half. This is the
origin of (\ref{baruch}, \ref{mono}); see Fig.~\ref{f1}. 

We stress that (\ref{baruch}, \ref{mono}) does not hold for a
random selection of the half of words. Expectedly, in that case the sign
of $r_{\rm min\,1}-r_{\rm min\,2}$ changes erratically from one text to
another, while $|r_{\rm min\,1}-r_{\rm min\,2}|$ is smaller than for the
natural division into the halves.  Also all other differences between
the two halves (discussed below) does not hold if the division is done
randomly. 

Table~\ref{tab0} shows that the parameter $c$ of Zipf's law
(\ref{barbos}) does not show any systematic behavior between the halves
\footnote{E.g.  $c_1=0.087<c_2=0.135$ for text OC [see
Table~\ref{tab1}], but $c_1=0.182>c_2=0.165$ for TS.}. Likewise, the
power-law exponent $\gamma$ in (\ref{barbos}, \ref{goodwin}) is
generally close to $1$ ($0.95\leq\gamma\leq 1.05$); it slightly differs
between the halves, but without a systematic trend. 

Once the range $1<r\leq r_{\rm min}$ contains many functional words,
we checked whether the two halves differ from each other by different
number of functional words. No systematic differences were found between
the halves.  We also divided the functional words into different
categories (conjunctions, pronouns, determiners) and checked each
category separately with the same negative result. 

\subsection{Rare words}

To describe the amount of rare words, we conventionally define a word as
rare, if it appears in the text at most 3 times. Denote by $h$ the
number of such words in a given text. We selected this threshold 3 so as
to get a robust comparison: since there is no a universal definition of
{\it word}, different methods of counting can lead to different results
\footnote{We found that the number of words that appear strictly once
does not show a regular behavior across of the halves.}; cf. Footnote
\ref{janmuller}. 

For all studied texts we observed [see Table~\ref{tab1}]:
\BEA
\label{cov}
h_1>h_2,
\EEA
where $h_1$ ($h_2$) is the number of words that appear at most 3 times in the
first (second) half of a given text. Eq.~(\ref{cov}) suggests that the first half uses more rare words, but
such a conclusion is incomplete, since the two halves have different numbers of distinct words. Denote them as 
$n_1$ and $n_2$, for the first and second half respectively. We saw a more
refined criterion that again holds for all studied texts [see Table~\ref{tab1}]:
\BEA
\label{boris}
h_1/n_1>h_2/n_2 
\EEA
Hence the first half has more rare words both in absolute and relative terms.
Both (\ref{cov}) and (\ref{boris}) are reproduced theoretically in section \ref{theor}. 

\section{Textual information}
\label{textual}

\subsection{Quantifiers of textual information}

We want to relate (\ref{baruch}--\ref{boris}) with textual features
developed qualitatively in linguistics; see \cite{hutchins} for a good
review, \cite{valgina} for a textbook and \cite{hasan} for a monograph
presentation. First of all, we recall that a text is a
hierarchic\footnote{Refs.~\cite{lrc_ebeling,lrc_eckmann,lrc_altmann}
recently mentioned hierarchic features of text in the context of
long-range correlations betweeen words and letters found in real texts.
} construct, i.e. it consists of several autonomous
levels\footnote{Autonomous means that features of a level are
constrainted, but cannot be completely deduced, from those of lower
levels.}: words, phrases, clauses, sentences, paragraphs
\footnote{Sometimes, they can coincide, i.e. a clause can coincide with
sentence, and there can be a one-paragraph sentence. } {\it etc}. 

The first level is that of words. Neglecting phenomena of synonymy and
homonymy (which are rare in English, but not at all rare e.g. in Chinese
\cite{epjb}), we can say that every word has several closely related
meanings (polysemy). Neglecting also the difference between polysemic
meanings, the number of independent meanings in a text can be estimated
via the number of different words. Further distinction between words of
the text can be made via their average length (in letters): content
words|which express specific meaning|are normally longer than functional words
that mostly serve for establishing grammatic connections \cite{zipf}.

The next level is that of clauses, which sometimes can (e.g. in simple
texts) coincide with a sentence. A clause joins several phrases, where
the (polysemic) meaning of separate words is clarified. Moreover, in
clauses there are new means of expressing meaning. Within many clauses
one can identify two types of phrases \cite{hutchins,hasan}: {\it
themes} present information that is already known from the preceding
text; {\it rhemes} could not be inferred by the reader and thus amount
to a new information \footnote{As an example, take the preceding
sentence: {\it Moreover, in clauses there are new means of expressing
meaning}. Here "moreover" is a textual theme, since it relates with
previous parts of the text; "in clauses" is a topical theme, and the
rest of the sentence (that contains the verb form) is the rheme. }.
Clauses joined in one sentence normally have the same theme
\footnote{Theme and rheme are different from, respectively, subject and
predicate that are grammatic constructs. 
But still in many English sentences, the subject and theme
coincide \cite{hasan}.}. Sometimes, clauses of the next sentence employ
the rheme of the previous sentence as their theme.  Hence the number of
clauses will indicate on the new information contained in the text
\cite{hutchins}.  We shall estimate the number of clauses in a text via
calculating the number of punctuation signs in the text \footnote{This
is not absolutely precise, because there are clauses that are connected
into a sentence without any punctuation sign; e.g. they can be connected
without comma, though colons, hyphens and semi-colons normally indicate
on a clause connection. Commas can be employed without
clause-connections, e.g. when listing items or emphasizing a part of a
sentence that is not a clause. But in other cases they do indicate on
the clause connection, e.g. when a comma is put before {\it but}, {\it
and}, or when marking an indirect speech.  Also recalling that we
compare two parts of the same text, we believe that in regular texts an
overall number of punctuation signs does correlate with the number of
clauses. }. 

A paragraph joins several sentences with closely related themes
\footnote{Instead of paragraph linguists frequently look at segments or
complex syntactic units \cite{valgina}. The main difference between such
constructs and the paragraph is that the latter is more dependent on a
specific writing style adopted by the text's author. For our purposes
this difference is not important, since we compare with each different
halves of the same text that are written by the same author, and
(normally) within the same style. }. Among these sentences there is one
(or few) that are autosemantic \cite{hutchins,valgina}, i.e.  they can
be taken out of the text, and they still retain their full meaning.
Frequently, the autosemantic sentence is the first one in the paragraph,
as it is the case with the present paragraph. The majority of sentences
in a paragraph are semantically dependent, i.e. they evolve around the
autosemantic sentences detailizing their meaning and/or providing
further information on them \cite{hutchins}.  For this higher level,
autosemantic (semantically dependent) sentences are analogues of theme
(rheme).  Thus a paragraph can also serve as a (higher-level) unit of
textual meaning.  Hence the number of paragraphs of a text is a relevant
descriptor of textual information. In this context note that there are
automatic routines that allow allow to fragment a given text over
topically homogeneous parts \cite{marti}.

Results presented below show that there is another (in a sense highest)
hierarchic level of the text organization that was so far not noted by
linguists \footnote{In this context we recall results from the language
processing literature showing that inter-sentence correlations between
words get stronger for sentences that are located deeper inside of a
paragraph \cite{charniak}.  Let $X_i^{[k]}$ be a random variable
denoting the $i$'th word that appears in the $k$ sentence of the
paragraph. For example, in the previous sentence, which is the second
sentence of the present paragraph, ``let'' is a value assumed by
$X_1^{[2]}$.  Values of $X_i^{[k]}$ are denoted as $x_i^{[k]}$.
We also define $X_{-i}^{[k]}\equiv (\, X_i^{[k]},X_{i-1}^{[k]},...,
X_1^{[k]} \,)$ for words of the same sentence, and let $Y^{[k]}$ to
define all words that appear in sentences $1,..., k-1$. Given a
sufficiently long text, one can estimate joint probabilities
$p(x_i^{[k]}, x_{-i}^{[k]}, y^{[k]})$ and hence calculate the
mutual-conditional information $I(X_i^{[k]}:Y^{[k]}\left | X_{-i}^{[k]}\right.) =
\sum_{x_i^{[k]}, x_{-i}^{[k]}, y^{[k]} } p\left(x_i^{[k]}, x_{-i}^{[k]},
y^{[k]}\right) \ln \frac{p(x_i^{[k]}, y^{[k]}| x_{-i}^{[k]})}{p(x_i^{[k]}\left|
x_{-i}^{[k]}\right.) p(y^{[k]}\left| x_{-i}^{[k]}\right.)}$, where 
$p(y^{[k]}\left| x_{-i}^{[k]}\right.)$ are conditional probabilities. Now
$I(X_i^{[k]}:Y^{[k]}\left | X_{-i}^{[k]}\right.)$
determines correlations
between $X_{i}^{[k]}$ and $Y^{[k]}$ given (i.e. conditioned upon)
$X_{-i}^{[k]}$.  Ref.~\cite{charniak} shows that
$I(X_i^{[k]}:Y^{[k]} \left| X_{-i}^{[k]}\right.)$ has a general trend of increasing with $k$ for a fixed
$i$. However, note that some of results in \cite{charniak} contradict to
literature \cite{cancho_debo}, e.g. the information constancy statement
that the conditional entropy $H(X_i^{[k]}\left|Y^{[k]} , X_{-i}^{[k]}\right.)$ is
constant as a function of $k$ contradicts to the verified Hilberg's
conjecture \cite{cancho_debo,debo}.  }. It amounts to differences
between the first and second halves of the text; in particular, the
first half contains more themes and more autosemantic sentences than the
second half. 

\subsection{Comparison between two halves of a text}

We found that above characteristics can distinguish different halves of
the same text, see Tables~\ref{tab0}, \ref{tab1} and \ref{tab2}.
However, for some of these distinctions there are exceptions | in
contrast to (\ref{baruch}, \ref{cov}, \ref{boris}), where we did not see
exceptions.  They are rare in the sense that for 10 studied texts we got
at most one exception. 

$\bullet$ The number of different words is larger in the first half
$n_1>n_2$; see Table~\ref{tab1} \footnote{One can anticipate here
possible relations with Herdan-Heap's law $n\sim N^{\beta}$
($\beta\approx 0.5$) that relates the number of different words $n$ in a
text with the overall number of words $N$. Finding a relation 
between Zipf's law and  Herdan-Heap's law was
attempted in \cite{kornai}.  }. 

$\bullet$ The total number of punctuation signs (where we
included full points, colons and semi-colons, commas, question marks and
exclamation points) is also larger in the first half: $S_1>S_2$; see
Table~\ref{tab1}. 

$\bullet$ The total number of letters employed is larger in the first
half: $L_1>L_2$; see Table~\ref{tab1}.  Hence, the average length of
words (in letters) is also larger in the first half. 

$\bullet$ We calculated the full distribution of sentences over the
length (measured in words): $\kappa_\alpha$ the fraction of sentences with
word-length $\alpha$ ($\sum_\alpha\kappa_\alpha=1$). Three specific
characteristics of this distribution are worth looking at: the average
$\overline{\alpha}$, dispersion $\overline{\Delta(\alpha^2)}$ and
entropy $\ve$:
\begin{gather}
\label{deviation}
\overline{\alpha}={\sum}_\alpha\kappa_\alpha\alpha, \qquad
\overline{\Delta(\alpha^2)}
={\sum}_\alpha\kappa_\alpha
(\alpha-\overline{\alpha})^2, \\
\label{entropy}
\ve=-{\sum}_\alpha\kappa_\alpha\ln\kappa_\alpha.
\end{gather}
Dispersion quantifies deviations from the average, while 
the entropy measures uncertainty, it is
minimal (maximal) for deterministic (homogeneous) probabilities. 

Now Table~\ref{tab2} shows that all these characteristics are smaller in the first half:
\BEA
\overline{\alpha}_1< \overline{\alpha}_2, \qquad
\overline{\Delta(\alpha^2)}_1<\overline{\Delta(\alpha^2)}_2, \qquad
\ve_1<\ve_2.
\EEA
In addition, Table~\ref{tab2} shows that the number of sentences in the
first are larger. This result is consistent with both
$\overline{\alpha}_1< \overline{\alpha}_2$ and $S_1>S_2$, taking into
account that the both halves have the same number of words. 

$\bullet$ The number of paragraphs is larger in the first half
(again with one exclusion): $\rho_1>\rho_2$; see Table~\ref{tab1}. 

\comment{Thus the first half has (approximately) more distinct words, longer
words, shorter (and more length-homogeneous) sentences, and more
paragraphs.}  

Altogether these features make intuitive sense, since they show that the
first half contains more themes and more autosemantic sequences
than the second half. 

Note that applying notions of information compression does not indicate
a robust difference between the two halves. Initially, the first half is
larger in bytes; which is natural given that its words are longer in
average; see Table~\ref{tab1}.  We compressed each half via zip,
Lempel-Ziv and several other standard routines. If the second half would
compress more than the first half (in absolute or relative units), we
would conclude that the second half has less information in the
Shannon's sense. However, we did not observe any indication that the
second half is compressed more than the first half.  

Table~\ref{tab1} shows that the discussed features (e.g. $n_1$ and $n_2$) are
normally closer to each for the halves of the same text, e.g. $|n_1({\rm
AR})-n_2({\rm AR})|\ll |n_1({\rm AR})-n_2({\rm TM})|$. Obviously, the
features of the two halves are close to each other. Similar points were
noted in \cite{baek} and stated as translation invariance of text
features. However, we stress that even small differences between the
halves can show systematic differences between them. 

\section{Spatial distribution of words}
\label{spatial}

\subsection{Spatial frequency versus ordinary frequency}
\label{spato}

Let us now turn to features that reflect the distribution of words along
the text. Studying this spatial distribution of words is traditional for
quantitative linguistics \cite{zipf,yngve}. More recently,
Refs.~\cite{ortuno,pury,carpena,zano} investigated the spatial
distribution of key-words versus functional words. The conclusion
reached is that key-words are distributed less homogeneously
\cite{ortuno,pury,carpena,zano}. Here we employ the spatial distribution
of words in the context of Zipf's law. 

Let $w_{[1]},...,w_{[\ell]}$ denote all occurrences of a word
$w$ along the text. Let $\zeta_{\,i\,j}$ denotes the number of words
(different from $w$) between $w_{[i]}$ and $w_{[j]}$. Define the 
average period $t(w)$ of this word $w$ via
\begin{equation}
\label{durnovo}
t(w)=\frac{1}{\ell-1}{\sum}_{l=1}^{\ell-1} \,(\zeta_{\,l\,\,l+1}+1).
\end{equation}
The averaging is conceptually meaningful only for sufficiently frequent 
words, though formally (\ref{durnovo}) is always well-defined. 
The inverse of this average quantity is the space-frequency $g(w)$:
\begin{equation}
\label{murnovo}
g(w)\equiv 1/t(w).
\end{equation}
Now if a sufficiently frequent word $w$ is distributed homogeneously,
then we expect $g(w)\approx\frac{\ell+1}{N+1}\approx f(w)=\ell/N$, where
$f(w)$ is the ordinary frequency, and where we assume that $N\gg 1$ and
$\ell\gg 1$. Hence the difference $g(w)-f(w)$ can tell how the
distribution of $w$ deviates from the homogeneous one. 

\begin{figure}
\centering
\includegraphics[width=0.5\textwidth]{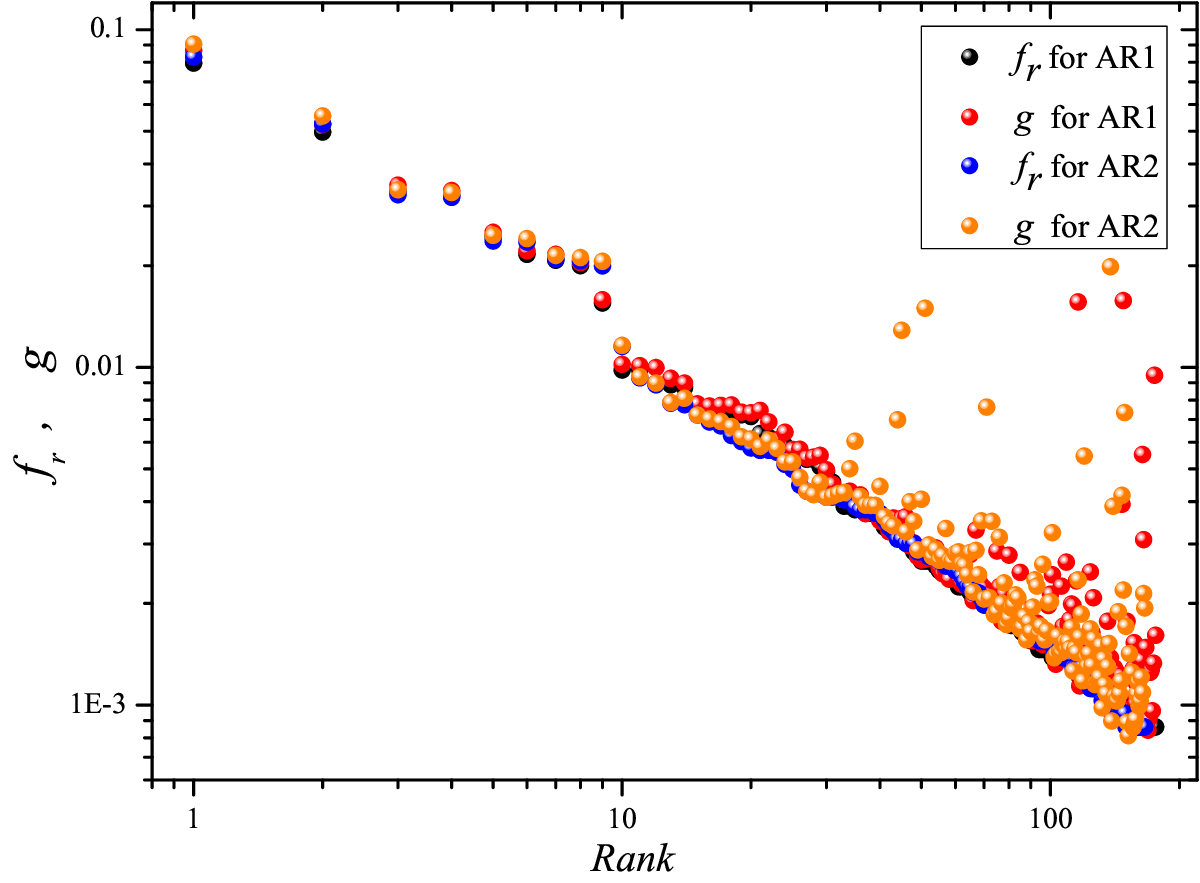}
\caption{ (Color online) Frequency $f_r$ vs. rank $r$ for the first (AR1)
and second (AR2) halves of the text AR; see Fig.~\ref{f1} and Table~\ref{tab1}. For each (ranked) 
word we also show its space-frequency (\ref{durnovo}). The shown ranks correspond to 
$r<r_{\rm max}$; cf.~Fig.~\ref{f1}.}
\vspace{0.25cm}
\label{f2}
\end{figure}

Fig.~\ref{f2} shows the ranked frequencies for the two halves of the
text AR.  For each word we show its (average) space-frequency $g(w)$
from (\ref{murnovo}) together with the (ordinary) frequency (\ref{w1}).
We stress that in Fig.~\ref{f2} the rank is defined via the ordinary
frequency ordering. Note that $g(w)$ roughly follows the general
pattern of Zipf's law. 

We see that for frequent words of each text (i.e.  AR1 or AR2) the
difference $|g(w)-f(w)|$ is small. This range of frequent words includes
the initial part of the Zipfian range $r_{\rm min}\leq r\leq r_{\rm
max}$; cf. Fig.~\ref{f2} with Fig.~\ref{f1}. It also includes all words
with ranks $1\leq r\leq r_{\rm min}$, i.e. those frequent words that do
not hold Zipf's law. Hence frequent words are distributed more
homogeneously \cite{zano}. 

But $|g(w)-f(w)|$ starts to grow already in the initial part of the
Zipfian range $r_{\rm min}<r<r_{\rm max}$. Fig.~\ref{f2} shows that this
growth is larger for the second half of AR. To quantify this point, we
defined normalized frequencies for the words in the Zipfian range:
\begin{equation}
\label{brutos}
\w{ g}(w_r)\equiv \frac{g(w_r)}{\sum_{r'=r_{\rm min}}^{r_{\rm max}}g(w_{r'})  },~~
\w{f}(w_r)\equiv \frac{f(w_r)}{\sum_{r'=r_{\rm min}}^{r_{\rm max}}f(w_{r'})  }.
\end{equation}
Once these frequencies are properly normalized within the Zipfian range, we can
quantify the distance between them via one of standard definitions of probability
distances. We choose the variational distance:
\begin{equation}
\mu={{\sum}_{r=r_{\rm min}}^{r_{\rm max}}|\w{f}(w_{r})-\w{g}(w_{r})|  }.
\end{equation}
Now Table~\ref{tab1} shows that for {\it all} studied texts there is a difference between the halves:
\begin{equation}
\label{conon}
\mu_2>\mu_1,
\end{equation}
i.e. in the first half the words of the Zipfian domain are distributed more homogeneously.

Note that the restriction to the Zipfian range $r_{\rm min}<r<r_{\rm
max}$ in (\ref{brutos}) is crucial for the validity of (\ref{conon}).
For instance, if we extend the definition to ranks $1\leq r \leq r_{\rm
max}$ (i.e. ${\sum}_{r'=r_{\rm min}}^{r_{\rm max}}\to {\sum}_{r'=1}^{r_{\rm
max}}$ everywhere) relation (\ref{conon}) does not hold anymore, i.e.
the sign of $\mu_2-\mu_1$ changes erratically from one text to another. 
This is an important point, because so far the Zipfian range 
$r_{\rm min}<r<r_{\rm max}$ was defined via strict, but still conventional 
criteria (\ref{dard}). Eq.~(\ref{conon}) shows that this definition does
capture an important feature of real texts. 

\subsection{Yule's constant}
\label{yu}

The above result|i.e. words in the Zipfian range are distributed more
homogeneously in the first half than in the second half|needs
corroboration. To this end, we looked at the Yule's constant \cite{baa}.
Define $V_m$ to be the number of words that (in a fixed text) appear $m$
times. We get two obvious features:
\begin{equation}
\label{kush}
{\sum}_{m=1}^{f_1 N}\, V_m=n, ~~{\sum}_{m=1}^{f_1N} \, mV_m=N, 
\end{equation}
where $n$ and $N$ are, respectively, the number of different words and the number of 
all words, and $f_1N$ is the number of times the most frequent word appears in the text.
Note that for a sufficiently small $m$, $V_m$ is either zero or one. For instance,
in the first half AR1 of AR, $V_{f_1N}=V_{929}=1$, $V_{928\geq m\geq 575}=0$, $V_{574}=1$,
$V_{573\geq  m\geq 388}=0$, $V_{387}=1$ {\it etc}. 

Take a word $w$ that appears $m$ times in a text with length $N$. Now
$\frac{m}{N}$ is the probability that a randomly taken word in the text
will be $w$. Likewise, $\frac{m(m-1)}{N(N-1)}$ is the probability that
the second randomly taken word in the text will be again $w$. Both probabilities
refer to a word $w$ that appears $m$ times. The probability to take
such a word among $n$ distinct words of the text is $\frac{V_m}{n}$ 
\footnote{Hence one can define the entropy $-\sum_m\frac{V_m}{n}\ln \frac{V_m}{n}$ that
characterizes the inhomogeneity of distribution of distinct words. In
contrast to the Yule's constant, the difference of this entropy
calculated between two halves of the same text changes erratically from
one text to another. This entropy was employed in \cite{cohen_fractal} for distinguishing between
natural and artificial texts.}.
Thus the average $\frac{1}{n}{\sum}_{m=1}^{f_1 N}\left[
V_m\frac{m(m-1)}{N(N-1)}\right]$ is a measure of repetitiveness of
words. The Yule's constant $K$ employs this quantity without the factor
$\frac{1}{n}$, since it wants to have something weakly dependent on $N$
\cite{baa}.  For us this feature is not important, since we compare the
halves of a text.  Following the tradition, we also omit the factor
$\frac{1}{n}$, but we stress that including it does not change our
conclusions. Using (\ref{kush}) and $N\simeq N-1\gg 1$, the Yule's
constant reads
\cite{baa}
\begin{equation}
\label{yule}
K=10^2\left[-\frac{1}{N}+
{\sum}_{m=1}^{f_1 N}\, V_m\,\frac{m^2}{N^2}\right],
\end{equation}
where $10^2$ is a conventional factor we applied to keep $K={\cal
O}(1)$; see Table~\ref{tab1}. 

We employ $K$ for comparing two halves of the same text with respect of
the repetitiveness of words. Table~\ref{tab1} shows that apart of one
exclusion we get $K_2>K_1$. In this sense words in the second half
repeat more frequently. This is consistent with previous findings, i.e.
$n_1>n_2$ (the first half has more different words) and that the first
half is fragmented more in the sense of a large number of paragraphs and
more inhomogeneous distribution of the sentence length. It is also
consistent with the fact established in section \ref{spato}, {\it viz.}
frequent words are distributed more homogeneously in the first part. 

\section{Theoretical description}
\label{theor}

\subsection{Remainder of the model}

Below we show that a statistical physics theory of Zipf's law
proposed in \cite{pre} can describe some of above effects. We stress
that this description is {\it incomplete}, e.g. the theory cannot
explain why specifically in the first half the applicability range of
Zipf's is larger. This is not surprising, since the theory is based
on purely statistical mechanisms of text-generation, i.e. it does not
account for semantic issues. But the theory confirms (\ref{cov},
\ref{boris}) and predicts new and useful relations that hold for halves.
The theory starts with the following assumptions. 

\comment{ The theory confirms the intuitive expectation about the
difference between the Zipfian and hapax legomena range: in the
first case the probability of a word is equal to its frequency
(frequent words). In the hapax legomena range, both the
probability and frequency are small and different from each other.}


$\bullet$ 
Given $n$ different words $\{w_k\}_{k=1}^n$, the joint probability
for $w_k$ to occur $\nu_k\geq 0$ times in a text $T$ is assumed to
be multinomial (the {\it bag-of-words model} \cite{madsen}) 
\BEA \label{multinom} \pi[\nunu
|\ff]=\frac{N!\,\theta_1^{\nu_1}...\theta_n^{\nu_n}}{\nu_1!...\nu_n!},
~\nunu = \{\nu_k\}_{k=1}^{n}, ~~ \ff=\{\theta_k\}_{k=1}^{n},
\EEA where $N=\sum_{k=1}^n\nu_k$ is the length of the text
(overall number of words), $\nu_k$ is the number of occurrences of
$w_k$, and $\theta_k$ is the probability of $w_k$.
According to (\ref{multinom}) the text is regarded to be a
sample of word realizations drawn independently with probabilities
$\theta_k$.

Eq.~(\ref{multinom}) is incomplete, because it implies that each
word has the same probability for different texts. In contrast, the same
words do {\it not} occur with same frequencies in different texts. 

$\bullet$ To improve this point we make $\ff$ a random vector with a
text-dependent density $P(\ff|T)$ \cite{hof,lazar}. With this assumption
the variation of the word frequencies from one text to another will be
explained by the randomness of the word probabilities.  Since $\ff$ was
introduced to explain the relation of $T$ with $\nunu$, it is natural to
assume that the triple $(T,\ff,\nunu)$ form a Markov chain: the text $T$
influences the observed $\nunu$ only via $\ff$. Then the probability
$p(\nunu|T)$ of $\nunu$ in a given text $T$ reads 
\BEA
p(\nunu|T)={\int}\d \ff \, \pi[\nunu |\ff]\, P(\ff |T).  \label{bay}
\EEA
Physically, $\ff$ refers to annealed disorder. 

$\bullet$ The text-conditioned density $P(\ff |T)$ is generated
from a prior density $P(\ff)$ via conditioning on the ordering of
${\bf w}=\{w_k\}_{k=1}^n$ in $T$: \BEA \label{chua} P(\ff |T)=
P(\ff)\,\chi_T(\ff,{\bf w})\left/ {\int}\d \ff' \, P(\ff')\,
  \chi_T(\ff',{\bf w})\right. .
\EEA If different words of $T$ are ordered as $(w_1,...,w_n)$
with respect to the decreasing frequency of their occurrence in
$T$ (i.e. $w_1$ is more frequent than $w_2$), then
$\chi_T(\ff,\bf{w})=1$ if $\theta_1\geq...\geq\theta_n$, and
$\chi_T(\ff,\bf{w})=0$ otherwise.

$\bullet$ Next, we assume what physically amounts to an ideal gas: the
probabilities $\theta_k$ are distributed identically and the dependence
among them is due to $\sum_{k=1}^n\theta_k=1$ only:
\BEA \label{g1} 
&& P(\ff)
\propto
\delta\left({\sum}_{k=1}^n\theta_k-1\right)\,
{\prod}_{k=1}^n Y(\theta_k),\\
&& Y(\theta_k)=({n}^{-1}{c}+\theta_k)^{-2},
\label{toki}
\EEA 
where $\delta(x)$ is the delta function and the normalization ensuring
$\int_0^\infty\prod_{k=1}^n \d \theta_k\, P(\ff)=1$ is omitted.
Eq.~(\ref{toki}) is a postulate that Ref.~\cite{pre} motivated via
relating it the mental lexicon (store of words) of the author. The
factor $c/n$ in (\ref{toki}) is necessary, since $P(\ff)$ is not
normalizable for $c/n=0$.  Here $c$ will be related to the prefactor of
Zipf's law (\ref{barbos}). 

An important feature of (\ref{toki}) is that the probability density
$Y(t_k)$ of the inverse frequency (or period) $t_k\equiv 1/\theta_k$
has the same shape as (\ref{toki}):
\BEA
\label{koki}
Y(t_k)
\propto ({c}^{-1}{n}+t_k)^{-2}.
\EEA
The meaning of (\ref{toki}, \ref{koki}) is that the frequency and
period|i.e., word choosing and period choosing|have the same distribution; cf. with section \ref{spato},
where we saw that the inverse period and the ranked frequency roughly
follow the same relation \footnote{This feature holds for all densities
$Y(\theta)\propto \left(\left[\frac{c}{n}\right]^\beta
+\theta^\beta\right)^\alpha$ with $\alpha\beta =-2$. We have chosen
(\ref{toki}), where $\beta=1$  and $\alpha=-2$, since it already 
provides a good fit to empiric data. This feature holds as well for 
$Y(\theta)\propto1/\theta$, which is however not normalizable, and hence cannot be employed.  }. 

Now note the following feature of real texts 
\BEA
\label{muftar}
n^3\gg N\gg n\gg 1,
\EEA
where $n$ is the number of
different words, while $N$ is the total number of words in the
text. Eq.~(\ref{muftar}) is verified for all texts we studied.

Eqs.~(\ref{multinom}--\ref{muftar}) allow to reach the final outcome of the theory: the probability
$p_r(\nu|T)$ of the word with the rank $r$ to
appear $\nu$ times in a text $T$ \cite{pre}: \BEA \label{g00}
p_r(\nu|T)=\frac{N!}{\nu!(N-\nu)!} \phi_r^{\nu}(1-\phi_r)^{N-\nu},
\EEA where the effective probability $\phi_r$ is found from
\BEA
\label{g6} \phi_r=cr^{-1}-cn^{-1}. 
\EEA 
If $\phi_r$ is sufficiently large (which is the case in the Zipfian
domain), $\phi_r N\gg 1$, the word with rank $r$ appears in the text
many times and its frequency $\nu\equiv f_rN$ is close to its maximally
probable value $\phi_rN$; see (\ref{g00}). Hence the frequency $f_r$ can
be obtained via the probability $\phi_r$. Then (\ref{g6}) is Zipf's
law generalized by the factor $n^{-1}$ at high ranks $r$. This cut-off
factor ensures faster [than $r^{-1}$] decay of $f_r$ for large $r$. In the
Zipfian range $cr^{-1}\gg cn^{-1}$ and (\ref{g6}) reverts to Zipf's law. 

\comment{
is determined 
from solving the following equation
\BEA \label{grm} \frac{r}{n}=ce^{-n
\phi_r\zmu}/(c+n \phi_r),~~
\zmu\equiv c^{-1}\,e^{-\gamma_{\rm E}-\frac{1+c}{c}}.
\EEA
This is the case in the Zipfian domain, since according to our empirical results \cite{pre} 
$\frac{1}{n}\lesssim f_r$ for $r\leq r_{\rm
max}$, and|upon identifying $\phi_r=f_r$|the above condition
$\phi_r N\gg 1$ is ensured by $N/n\gg 1$.
Let us return to (\ref{grm}). For $r>r_{\rm min}$,
$\phi_rn\zmu=f_rn\zmu<0.04\ll 1 $. We get from (\ref{grm}): }

According to (\ref{g00}), the probability $\phi_r$ is small for $r\gg
r_{\rm max}$ and hence the occurrence number $\nu\equiv f_rN$ of the
word with the rank $r$ is a small integer (e.g.  1 or 2) that cannot be
approximated by a continuous function of $r$.  To describe this hapax
legomena range, define $r_k$ as the rank, when $\nu\equiv f_rN$ jumps
from integer $k$ to $k+1$ (hence the number of words that appear $k+1$
times is $r_{k}-r_{k+1}$).  Since $\phi_r$ reproduces well the trend of
$f_r$ even for $r>r_{\rm max}$, $r_k$ can be theoretically predicted
from (\ref{g6}) by equating its left-hand-side to $k/N$: \BEA
\hat{r}_k=\left[\frac{k}{Nc}+\frac{1}{n}\right]^{-1}, \qquad k=0,1,2,...
\label{hapo} \EEA Eq.~(\ref{hapo}) is exact for $k=0$, and agrees with
$r_k$ for $k\geq 1$ with a small relative error \cite{pre}. Hence
a single formalism describes both Zipf's law for short texts and the
hapax legomena range. For describing the hapax legomena
no new parameters are needed; it is based on the same parameters $N,\,
n,\, c$ that appear in Zipf's law. 

\subsection{Applying the model to the halves}

The number of words $V_m$ that appear $m$ times is expressed as 
$V_m=r_{m-1}-r_{m}$ in terms of the above parameter $r_m$. Using
(\ref{hapo}) we provide a theoretical estimate for this quantity
times
\begin{eqnarray}
\hat V_m &=&\hat r_{m-1}-\hat r_{m}\nonumber\\
&=&\frac{Nc}{\left[m-1+\frac{Nc}{n}\right]\, \left[m+\frac{Nc}{n}\right] }, \qquad m\geq 1.
  \label{vem}
\end{eqnarray}
Eq.~(\ref{vem}) applies well for sufficiently small values of $m$ \cite{pre}, and
replaces the so called Lotka-Zipf's law $V_m\propto
m^{-2}$ \cite{zipf_2}, which is much less precise. 

Now (\ref{cov}, \ref{boris}) are reproduced from (\ref{vem}) upon
putting there the fitted values of $c_1$ and $c_2$ (from
Table~\ref{tab2}), the total number of words $N_1$ and $N_2$, and the
number of different words $n_1$ and $n_2$ (from Table~\ref{tab1}) for
each half. 

Note from (\ref{g1}) that the parameter $c/n$ characterizes the width of
the prior word density, i.e. a smaller $c/n$ means that more
low-probable words are involved. An important observation is that for
all texts we analyzed this ratio is smaller in the first half [see
Table~\ref{tab2}]:
\BEA
\label{bri}
c_1/n_1<c_2/n_2.
\EEA
This result is consistent with the above model [see (\ref{toki})], since it heuristically
predicts that the first half involves more rare words. 

\comment{Also, a smaller regularizing factor in (\ref{g1}) implies that the
emergent power law will have a larger applicability domain, and this is
what we observed empirically. }

Now (\ref{g6}) and (\ref{bri}) imply that the theoretical prediction for
Zipf's law is more precise for the first half than for the second
half. However, we stress that the presented theory|given the values of
$c_k$ and $n_k$ ($k=1,2$)|does not reproduce precisely the empiric
values $r_{\rm min}$ and $r_{\rm max}$ for the Zipfian range.

\comment{
In more detail, this point can be verified
from rewriting (\ref{grm}) as (upon identifying $\phi_r$ with $f_r$)
\BEA
\label{panev}
\ln\left[cr^{-1}/f_r  \right]=nf_r\mu+\ln(1+\frac{c}{nf_r}).
\EEA
Hence the right-hand-side of (\ref{panev}) characterizes the deviation of the theoretically
predicted frequency $f_r$ from Zipf's law (\ref{barbos}, \ref{goodwin}). We checked for all 10 texts that 
for sufficiently large ranks (typically $r>10-25$), (\ref{panev})|supplemented by empiric values 
of $c_k$ and $n_k$ ($k=1,2$)|predicts a larger deviation from Zipf's law in the second half,
as compared with the first half.}

\section{Summary}

Our aim was to relate Zipf's law (\ref{w1}, \ref{barbos}, \ref{goodwin}) to
meaning-carrying features of the text. First we had to understand in
which specific sense this law is a statistical regularity (roughly akin
to the law of large numbers). To this end, we studied the validity range
of the law upon taking together two different texts; see section
\ref{mixing}. The validity range does increase, but this increase
happens mostly due to low-frequency words.  Next, in section \ref{divi}
each text was divided into two halves. This allows to uncover hidden
relations, since various confounding variables (genre, style, the
author's vocabulary {\it etc}) are the same in both halves.  For the
first half, Zipf's law applies from smaller ranks, and its validity
range covers more frequent words. 

On the other hand, we uncovered several textual features that are
different between the halves; see Table~\ref{tab0}. In section
\ref{textual} we reviewed several basic notions of text linguistics,
e.g. theme vs. rheme in clauses, and autosemantic sentences of
paragraphs; see \cite{hutchins} for a more detailed review.  We argued
that quantitative differences between the halves (shown on
Table~\ref{tab0}) imply that the first half of the text has more
thematic information than the second half. We suggest that Zipf's
law is related to the presence and amount of this information.
\comment{One possible interpretation of (\ref{baruch}, \ref{mono}) and
(\ref{conon}) is that the first half consists of a larger number of
sub-texts that (upon mixing) leads to better applicability of Zipf's
law.  This is not correct: mixing mostly increases $r_{\rm max}$ and
does not change the $r_{\rm min}$ in any systematic way; cf. sections
\ref{mixing} and \ref{dividivi}.}

For describing the above relations of Zipf's law we need new models,
since the existing statistical and optimization models do not account
for meaning-carrying features of texts; see section \ref{intro} for a
review.  However, the statistical model developed in \cite{pre} can be
applied for confirming some of the observed empiric relations and for
predicting new ones; see (\ref{bri}). This fact was demonstrated in
section \ref{theor}. 

Relations discussed in the first group of Table~\ref{tab0} only require
that the text consists of well-defined (but possibly unknown) words, no
futher structure of the text is needed. These are: the minimal rank (and
the maximal frequency) of Zipf's law, homogeneity of the spatial
distribution for the Zipfian words, the number of different words, the
number of rare words, the normalized prefactor of Zipf's law. These
features can be employed for finding out whether a sequence of word-like
symbols (written in unknown system) constitutes a text. Several other
regularities are known in literature that distinguish between a text and
a random string of words. Ref.~\cite{lande} found that texts are
compressed better in their natural ordering of words than after
sufficiently many random permutations of words. Ref.~\cite{debo} argues
that the scaling behavior of the mutual information between different
long block of a text can indicate on its difference from the random
text. 

Remaining relations we found|see the second part of Table~\ref{tab0}|do
require that finer-grained text structures are known and available, e.g.
that words consist of letters (which is not true in non-alphabetic
writing systems) or that the text is fractioned into sentences and
paragraphs (not the case in cryptic texts) {\it etc}. These features are
important, because they uncover a textual structure that goes
well-beyond paragraphs and chapters. Though they cannot be directly
employed for the above task of text recognition, there are possible
relations between some features from the first versus second group. We
submit them for future consideration. 

-- Recall the relation between rare words and frequent words that goes
via sentences: each sentence normally contains both frequent words and
rare words; cf. Footnote~\ref{lato}. Hence we expect a relation between
larger number of sentences (in the first half) and the following two
facts. First, Zipf's law starts from smaller ranks and larger
frequency; see (\ref{baruch}, \ref{mono}). Second, there are more rare
words in the first half; cf.~(\ref{cov}, \ref{boris}). 

-- More punctuation signs in the first half obviously correlate with
shorter sentences; see Table~\ref{tab0}. We also expect that longer
sentences in the second half correlate with a larger word
repetitiveness, as quantified by the Yule's constant (\ref{yule}). Also,
there can be a direct connection between spatial hamogeneity of Zipfian
words seen in section \ref{spato}, and the number of sentences. 

-- We anticipate a relation between smaller dispersion and entropy of
the sentence distribution (from one hand) and the larger number of
paragraphs (on the other hand). 

Clarifying these points may uncover further connections between
semantics and statistics, and also help to establish a text as one of
the basic subjects of complex systems theory; see
\cite{lrc_ebeling,lrc_eckmann,lrc_altmann,lande,debo,cancho_debo} in
this context. 

\comment{
\section{Conclusion (to be written)}

--  Discuss Ref.~\cite{zano} with more detail.

-- Stress novelties from the linguistic viewpoint.

-- Mention internal speech (Luria etc).

-- Read the book by Moscalchuk.

-- Read the article by Petrov.

-- Read other two papers by Hutchins.

-- Read your own old notebook with more detail. 

-- What is the empiric meaning of $c/n$, i.e. its meaning independently
from the theoretic model? 

}

\section*{Acknowledgements}

W. Deng was partially supported by the Program of Introducing Talents of
Discipline to Universities under grant no. B08033, and National Natural
Science Foundation of China (Grant No. 11505071). A.E. Allahverdyan
was supported by SCS of Armenia, grant 18RF-015.

\clearpage 

\section*{\Large Appendices}

\section*{Appendix I: Numeric data}
\label{num}

\begin{table}[h!] 
\centering 
\tabcolsep0.038in \arrayrulewidth0.3pt
\renewcommand{\arraystretch}{0.9}
\begin{tabular}{lccccccccccccccccc}
\hline\hline
Texts & $N$ &           $n_1$ & $n_2$ & $r_{{\rm min}\,1} $& $r_{{\rm min}\,2} $ & $S_1$ & $S_2$& $L_1$ & $L_2$ & $h_1$ & $h_2$ & $K_1$ & $K_2$ & $\mu_1$ & $\mu_2$ & $\rho_1$ & $\rho_2$   \\
\hline
AR & $2\times 11612$ &  2012 &  1958  & 18 & 33 & 1431 & 1399   &51305 &50889 &1600 &1520 &1.471 & 1.555 &0.270 &0.39 &157 &112 \\
DL & $2\times 14061$ &  2461 &  2421  & 13 & 22 & 2001 & 1812  &58975 &58124  &1930 &1895 &0.834 & 0.882 &0.066 &0.074 &334 &290 \\
TF & $2\times 13723$ &\u{2445}&\u{2725}  & 14 & 30 & 1554 & 1531 &63742 &62160&1961 &1606 &1.384 &1.412  &0.125 &0.130 &49 & 34 \\
TM & $2\times 16387$ &  3222 &  3111  & 17 & 44 & 2345 & 2162   &71383 &69963 &2647 &2541 &1.090 &1.288  &0.190 &0.20  & 202 & 144 \\
DA & $2\times 12411$ &  2490 &  2300  & 11 & 23 & \u{1634} & \u{1773}   &51806 &50949 &1983 &1789 &\u{0.948} &\u{0.911} &0.090 & 0.096 &279 & 252 \\
DM & $2\times 11511$ &  1972 &  1878  & 17 & 23 & 1219 & 1099   &52097 &50698 &1526 &1451 &1.250 &1.262 &0.078 &0.29 &49 & 23 \\
OC & $2\times 12756$ &  1987 &  1683  & 24 & 33 & 1602  & 1416   &\u{60415} &\u{60523}&1458 &1225  &1.575 &1.804 &0.16 & 0.19 &185 &175\\
TO & $2\times 12003$ &  2214 &  2183  & 22 & 34 & 1593  & 1465  &56015 &55653 &1745 &1733  &1.161 &1.209 &0.15 & 0.16 & 86 & 72\\
TS & $2\times 18069$ &  2970 &  2744  & 32 & 40 & 3524  & 3035  &75836 &74818 &2318 &2110 &0.719 &0.822 & 0.085 & 0.12 & 1170 & 826 \\
WW & $2\times 12217$ &  2868 &  2674  & 14 & 24 & 2040  & 1861  &52204 &51753 &2366 &2155 &0.899 &1.016 & 0.1 & 0.11 & \u{475} & \u{534}\\
\hline
\end{tabular}
\caption{ \label{tab1}Parameters of 10 texts: {\it
The Age of Reason} (AR) by T. Paine, 1794 (the major source of British
deism). {\it Thoughts on the Funding System and its Effects} (TF) by P.
Ravenstone, 1824 (economics).  {\it Dream Lover} (DL) by J. MacIntyre,
1987 (romance novella). {\it Time Machine} (TM), by H. G. Wells, 1895
(science fiction).  {\it Dawn of Avalon} (DA), by Anna Elliott, 2010
(historical fantasy).  {\it Discourse on the Method of Rightly
Conducting the Reason and Seeking for Truth in the Sciences} (DM), by
Rene Descartes, 1637 (natural philosophy).  {\it Treatise on the Origin
of Language} (TO), by Johann Gottfried Herder, 1772 (historical
linguistics). {\it Tidal Swans}, by James Welsh (TS), 2011 (romance
novella). {\it Overproduction and Crises}, by Karl Rodbertus (OC), 1898
(economics).  {\it Whirl of the Wheel} (WW), by Catherine Condie, 2009
(science fiction).\\ Total number of words is $N$, the number of
different words is $n$, $r_{\rm min}$ is the lower rank of the Zipfian domain, $S$
is the number of punctuation signs, $L$ is the number of letters, 
$h$ is the number of words that appear less then 4 times, $K$ is the Yule's constant, 
$\mu$ is the normalized distance, and $\rho$ is the number of paragraphs. 
The lower indices, e.g. in $S_1$ and $S_2$ 
refer to the first and second halves, respectively. Underlined pairs are atypical 
with respect to the halves, e.g. everywhere besides TF we get $n_1>n_2$.}
\end{table}

\comment{
\begin{table*}[h!]
\begin{center}
\tabcolsep0.05in \arrayrulewidth0.3pt
\renewcommand{\arraystretch}{0.4}
\caption{
Parameters of the texts defined in Table~\ref{tab1}: the number of different words $n$,
the lower bound $r_{min}$ and the upper bound $r_{max}$ of the ranks in the Zipfian
domain, the fitted values of $c$ and $\gamma$, and the difference $d$
between the empirical frequency of the Zipfian domain and its value
according to Zipf's law: $d=\sum_{k=r_{\rm min}}^{r_{\rm max}}
(ck^{-\gamma}-f_k)$. TF and TM means joining the texts TF and TM.}
\begin{tabular}{cccccccc}
\hline\hline
Texts &  $n$ &$r_{min} $& $r_{max}$ & $c$ & $\gamma$ & $\vert$$d$$\vert$   \\
\hline
TF &         2067 &  36   & 371 & 0.168 & 1.032 & 0.00333\\
TM &         2612 &  42   & 332 & 0.166 & 1.041 & 0.01004\\
AR &         1706 &  32   & 339 & 0.178 & 1.038 & 0.00048\\
DL&          1748 &  34   & 230 & 0.192 & 1.039 & 0.02145\\
TF and TM &  3408 &  30   & 602 & 0.139 & 1.013 & 0.02091\\
TF and AR &  2656 &  33   & 628 & 0.138 & 0.998 & 0.00239\\
TF and DL &  2877 &  28   & 527 & 0.162 & 1.014 & 0.01490\\
TM and AR &  3184 &  43   & 592 & 0.157 & 1.021 & 0.00491\\
TM and DL &  3154 &  45   & 493 & 0.161 & 1.023 & 0.01211\\
AR and DL &  2550 &  38   & 496 & 0.165 & 1.012 & 0.00947\\
Four texts&  4047 &  39   & 927 & 0.158 &  1.015& 0.00187\\
\hline
\label{tab33}
\end{tabular}
\end{center}
\end{table*}
}

\begin{table}
\centering 
\tabcolsep0.08in \arrayrulewidth0.3pt
\renewcommand{\arraystretch}{0.9}
\begin{tabular}{cccccccc}
\hline\hline
Texts &  $n$ &$r_{min} $& $r_{max}$ & $c$ & $\gamma$ & $\vert$$d$$\vert$   \\
\hline
AR &          3021 &  21  & 219 & 0.139& 1.005 & 0.00033\\
DL &          3610 &  30  & 315 & 0.163 & 1.038 & 0.00013\\
TF &          3896 &  23  & 287 & 0.185& 1.079 & 0.00375\\
TM &          4692 &  41  & 279 & 0.157 & 1.039 & 0.00053\\
AR and DL &   5542 &  27  & 518 & 0.171 & 1.059 & 0.00219\\
AR and TF &   5586 &  36  & 643 & 0.158 & 1.054 & 0.00196\\
AR and TM &   6451 &  34  & 586 & 0.155& 1.047 & 0.00392\\
DL and TF &   6288 &  31  & 482 & 0.178 & 1.075 & 0.00267\\
DL and TM &   6668 &  39  & 473 & 0.156& 1.051 & 0.00159\\
TF and TM &   7071 &  38  & 550 & 0.180& 1.082 & 0.00256\\
Four texts&   9857 &  42  & 969 & 0.169 &  1.074& 0.00431\\
\hline
\end{tabular}
\caption{
\label{tab33}Parameters of the texts defined in Table~\ref{tab1}: the number of different words $n$,
the lower bound $r_{min}$ and the upper bound $r_{max}$ of the ranks in the Zipfian
domain, the fitted values of $c$ and $\gamma$, and the difference $d$
between the empirical frequency of the Zipfian domain and its value
according to Zipf's law: $d=\sum_{k=r_{\rm min}}^{r_{\rm max}}
(ck^{-\gamma}-f_k)$. AR and DL means joining the texts AR and DL.}
\end{table} 

\begin{table}
\centering 
\tabcolsep0.05in \arrayrulewidth0.3pt
\renewcommand{\arraystretch}{0.9}

\begin{tabular}{lcccccccccccccc}
\hline\hline
Texts & $B_1$ & $B_2$ & $\si_1$ & $\si_2$& $\overline{\alpha}_1$ & $\overline{\alpha}_2$ &$\ve_1$ & $\ve_2$ & $\overline{\Delta(\alpha^2)}_1$ & $\overline{\Delta(\alpha^2)}_2$ & $c_1$ & $c_2$ & $\delta$ & $\widetilde{\delta}$  \\
\hline
AR & 97792 & 96256 & 1407 & 1378 &8.316 & 8.505 &2.934 &2.978 &29.61 & 32.21 & 0.135 & 0.135 & 1.850 &3.409 \\
DL & 112640 & 109568 &1988 &1797 &7.513 & 8.196 &2.877 &2.961 &26.08 &31.25 &0.138 & 0.148 & 5.057 &3.888 \\
TF & 112640  & 112128 &1499 &1449 &9.247 &9.473 &2.988 &3.020 &29.39 & 30.46 &0.148 & 0.186 & 7.725 &4.371 \\
TM & 130560 & 126976 &2338 &2059 &7.373 & 7.964&2.832 &2.899 &24.09 &24.46 &0.107 & 0.105 & 0.542  &3.908 \\
DA & 103424 & 101376 &\u{1538} &\u{1697} &\u{8.144}& \u{7.340} &\u{2.967} &\u{2.846} &\u{35.43} &\u{26.37} &0.129 & 0.135 & 6.888 &5.858  \\
DM & 93696 & 90624 &1214 &1093 &9.512 & 10.563 &3.115 &3.223 & 43.27 & 54.60 &0.148 & 0.151 & 5.354 & 2.141\\
OC &117760 & 108544 &1588 &1398 &8.207 &9.401 &3.008 &3.151 &43.27 &60.54 &0.087  & 0.135 & 36.43 &\u{-0.466}\\
TO & 101 888 & 100352 &1480 &1368 &8.165 & 8.920 &0.000 &0.000 &38.11 &49.96 & 0.135  & 0.145& 5.447 &1.872  \\
TS & 154112 & 145408 &3345 &2864 &5.436 & 6.369&2.555 & 2.716 &15.51 &17.52 & 0.182  & 0.169& 0.309 &1.463 \\
WW & 117760 & 116224 &1799 & 1781 &6.913 &6.979 &2.770 &2.800 &30.08 &30.62 & 0.117  & 0.110& 0.342 &3.774 \\
\hline
\end{tabular}
\caption{\label{tab2} Parameters of 10 texts not
included in Table~\ref{tab1}. $B$ is the text size measured in bytes.
$\sigma$ is the number of sentences.  $\ve$ is the entropy of sentence
distribution over lengths (counted in words); see (\ref{entropy}).
$\overline{\alpha}$ and $\overline{\Delta(\alpha^2)}$ are respectively
the average length of sentences in words and the standard deviation of the
sentence distribution; see (\ref{deviation}).  $c$ is the fitted values
of the parameter in (\ref{barbos}), $\delta\equiv 10^6 (\frac{c_2}{n_2}
-\frac{c_1}{n_1})$, and $\widetilde{\delta}=10^3[\frac{c_1}{r_{\rm min 1}}
-\frac{c_1}{r_{\rm min 1}}]$. The lower indices, e.g. in $c_1$ and $c_2$ refer to
the first and second halves, respectively. Underlined pairs are atypical
with respect to the halves. Note that here all atypical cases relate to
one text DA.}
\end{table}

\clearpage

\section*{Appendix II: Linear fitting}
\label{linear}

For each text we extract the ordered frequencies of different words [the
number of different words is $n$; the overall number of words in a text
is $N$]: \BEA \label{001} \{f_r\}_{r=1}^{n}, ~~ f_1\geq ...\geq f_{n},
~~ {\sum}_{r=1}^{n} f_r =1. \EEA We should now see whether the data
$\{f_r\}_{r=1}^{n}$ fits to a power law: $\hat{f}_r=cr^{-\gamma}$.  We
represent the data as \BEA \{y_r(x_r)\}_{r=1}^{n}, ~~ y_r=\ln f_r, ~~
x_r=\ln r, \EEA and fit it to the linear form $\{\hat{y}_r=\ln c-\gamma
x_r\}_{r=1}^{n}$. Two unknowns $\ln c$ and $\gamma$ are obtained from
minimizing the sum of squared errors: \BEA S_{\rm err}={\sum}_{r=1}^{n}
(y_r-\hat{y}_r)^2. \EEA It is known since Gauss that this minimization
produces \BEA \label{laude} -\gamma^* = \frac{ \sum_{k=1}^n
(x_k-\overline{x})(y_k-\overline{y}) }{ \sum_{k=1}^n
(x_k-\overline{x})^2}, ~~ \ln c^*=\overline{y}+\gamma^*\overline{x},
\EEA where we defined \BEA \overline{y}\equiv \frac{1}{n}{\sum}_{k=1}^n
y_k, ~~ \overline{x}\equiv \frac{1}{n}{\sum}_{k=1}^n x_k. \EEA As a
measure of fitting quality one can take: \BEA {\rm
min}_{c,\gamma}[S_{\rm err}(c,\gamma)] =S_{\rm err}(c^*,\gamma^*)\equiv
S_{\rm err}^*. \EEA This is however not the only relevant quality
measure. Another (more global) aspect of this quality is the coefficient
of correlation between $\{y_r\}_{r=1}^{n}$ and
$\{\hat{y}_r\}_{r=1}^{n}$: \BEA R^2=\frac{ \left[\, \sum_{k=1}^n
(y_k-\bar{y}) (\hat{y}^*_k-\overline{\hat{y}^*})\, \right]^2
}{\sum_{k=1}^n (y_k-\bar{y})^2 \sum_{k=1}^n
(\hat{y}^*_k-\overline{\hat{y}^*})^2}, \EEA where \BEA
\hat{y}^*=\{\hat{y}^*_r=\ln c^*-\gamma^* x_r\}_{r=1}^{n}, ~~~
\overline{\hat{y}^*}\equiv \frac{1}{n}{\sum}_{k=1}^n \hat{y}^*_k.  \EEA
For the linear fitting (\ref{laude}) the squared correlation coefficient
is equal to the coefficient of determination, \BEA R^2= {\sum}_{k=1}^n
(\hat{y}^*_k-\overline{y})^2\left/ {\sum}_{k=1}^n (y_k-\overline{y})^2,
\right. \EEA the amount of variation in the data explained by the
fitting. Hence $S_{\rm err}^*\to 0$ and $R^2\to 1$ mean good fitting. 

\section*{Appendix III: Kolmogorov-Smirnov (KS) test}
\label{ks}

We wanted to have an alternative method for checking the quality of the
above least-square method. To this end we applied the Kolmogorov-Smirnov
(KS) test to our data on the word frequencies. The empiric results on
word frequencies $f_r$ in the Zipfian range $[r_{\rm min}, r_{\rm max}]$
are fit to a power law. With null hypothesis that empiric data follows
the numerical fittings, we calculated the maximum differences (test
statistics) $D$ and the corresponding p-values (between empiric data and
numerical fitting) in the KS tests. Here are typical numbers for 3
texts: $D({\rm TF})=0.0418$, $p({\rm TF})=0.865$; $D({\rm AR})=0.0564$,
$p({\rm AR})=0.624$; $D({\rm DL})=0.0451$, $p({\rm DL})=0.812$.  One
sees that all the test statistics $D$ are quite small, while the
p-values are {\it much larger} than 0.1. We conclude that from the
viewpoint of the KS test the numerical fittings and theoretical results
can be used to characterize the empiric data in the Zipfian range
reasonably well. 

\comment{
\section{Normalization constant of Zipf's law}
\label{c}

Note that since Zipf's law (\ref{barbos}) does not apply
  for all ranks, $c$ is not a normalization constant, i.e. its value
  is not fixed from the fact that the sum of probabilities should be
  equal to $1$. Still the normalization allows to put upper and lower
  bounds on $c$. First recall from Fig.~\ref{f1} that the Zipf law is an
  upper bound for the frequencies at all ranks. This holds generally
  \cite{wyllis,baa}. Then we get using the known formula for $\sum_{k=1}^nk^{-1}$:
\BEA c\sum_{k=1}^nk^{-1}=
  c\left[\gamma_{E}+\ln n+\frac{1}{2n}+ {\cal O}(\frac{1}{n^2})\right] >1,
\EEA 
where $\gamma_{\rm E}=0.55117$ is the Euler's constant. On the other
hand, within the applicability range of Zipf's law we have
\BEA c\sum_{k=r_{\rm min}}^{r_{max}}k^{-1}\simeq
  c(\gamma_{E}+\ln \frac{r_{\rm max}}{r_{\rm min}})<1. 
\EEA 
These two formulas bound $c$ from above and below; e.g. for the text TF [see
Table~\ref{tab1}] they produce: $0.1218<c<03436$. Hence the bounds do not explain
the fact that for real texts $c< 0.2$; see Table~II. 
}

\end{document}